\documentclass{article}
\usepackage{natbib} 
\usepackage{arxiv}
\usepackage{array}
\usepackage{multirow}
\usepackage[utf8]{inputenc} 
\usepackage[T1]{fontenc}    
\usepackage{hyperref}       
\usepackage{url}            
\usepackage{booktabs}       
\usepackage{amsfonts}       
\usepackage{nicefrac}       
\usepackage{microtype}      
\usepackage{subcaption}    
\usepackage{lipsum}		
\usepackage{graphicx}
\usepackage{doi}
\usepackage[table]{xcolor}
\usepackage{float}
\title{Stage-Specific Cancer Survival Prediction
Enriched by Explainable Machine Learning}

\author{ \href{https://orcid.org/0009-0004-3012-7763}{\includegraphics[scale=0.06]{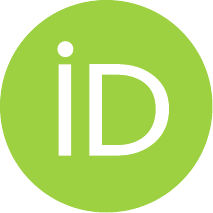}\hspace{1mm}Parisa Poorhasani}\\
	Department of Information Technology\\ Faculty of Science and Engineering\\ Åbo Akademi University\\ Turku, Finland\\
	\texttt{poorhasaniparisa@gmail.com} \\
	\And
	\href{https://orcid.org/0000-0001-8801-7250}{\includegraphics[scale=0.06]{orcid.pdf}\hspace{1mm}Bogdan Iancu}
    \thanks{Corresponding author.} \\
	Department of Information Technology\\ Faculty of Science and Engineering\\ Åbo Akademi University\\ Turku, Finland\\
	\texttt{bogdan.iancu@abo.fi} \\
}

\date{}

\renewcommand{\shorttitle}{Stage-Specific Cancer Survival Prediction
Enriched by Explainable Machine Learning}

\hypersetup{
pdftitle={Stage-Specific Cancer Survival Prediction
Enriched by Explainable Machine Learning},
pdfsubject={},
pdfauthor={Bogdan~Iancu, Parisa~Poorhasani},
pdfkeywords={First keyword, Second keyword, More},
}

\begin{document}
\maketitle

\begin{abstract}
Despite the fact that cancer survivability rates vary greatly between stages, traditional survival prediction models have frequently been trained and assessed using examples from all combined phases of the disease. This method may result in an overestimation of performance and ignore the stage-specific variations. Using the SEER dataset, we created and verified explainable machine learning (ML) models to predict stage-specific cancer survivability in colorectal, stomach, and liver cancers. ML-based cancer survival analysis has been a long-standing topic in the literature; however, studies involving the explainability and transparency of ML survivability models are limited.
Our use of explainability techniques, including SHapley Additive exPlanations (SHAP) and Local Interpretable Model-agnostic Explanations (LIME), enabled us to illustrate significant feature-cancer stage interactions that would have remained hidden in traditional black-box models. We identified how certain demographic and clinical variables influenced survival differently across cancer stages and types. These insights provide not only transparency but also clinical relevance, supporting personalized treatment planning. By focusing on stage-specific models, this study provides new insights into the most important factors at each stage of cancer, offering transparency and potential clinical relevance to support personalized treatment planning.	
\end{abstract}

\keywords{Cancer Survival Prediction \and Stage-specific survivability \and Explainable Machine Learning}

\section{Introduction}
{C}{ancer} is a major public health issue. Its global incidence is expected to show a progressive increase, primarily attributable to demographic aging, lifestyle-related risk factors, and exposure to environmental carcinogens. By 2040, an estimated 28 million additional cases and 16.2 million deaths worldwide are expected ~\cite{liu2024exploring,mattiuzzi2019current,tang2024long}. The rates of morbidity and mortality linked to cancer have led academics and bioinformaticians to develop advanced machine learning (ML) models designed to predict cancer survivability~\cite{el2024risk,buk2023machine}.

 Survival analysis, as a statistical model, is extensively utilized in clinical oncology to provide patients with prognostic information by determining the likelihood of survival beyond a specified duration~\cite{deepa2022systematic}. ML-based predictive models for cancer survivability are becoming increasingly important because of their potential to improve cancer care and management. These ML algorithms provide accurate predictions of patient outcomes by examining various types of data, including tumor features, treatment, and patient demographics. Currently, machine learning models for predicting cancer survival rates have increased significantly, with the objective of delivering more precise prognostic evaluations for individual patients~\cite{pour2018stage}.

However, accurate prediction of cancer survivability requires an understanding of how different stages of the disease influence patient outcomes, as each stage represents a different degree of tumor advancement and has different prognostic implications. In this study, we focused on three main stages classified by SEER~\cite{seer}: localized (confined to the primary organ), regional (extension to adjacent lymph nodes, tissues, or organs), and distant (metastasis to remote sites). This stage-specific stratified approach allows for a more refined modeling of survival outcomes.
\begin{figure*}[!htb]
    \centering
    \includegraphics[width=\columnwidth]{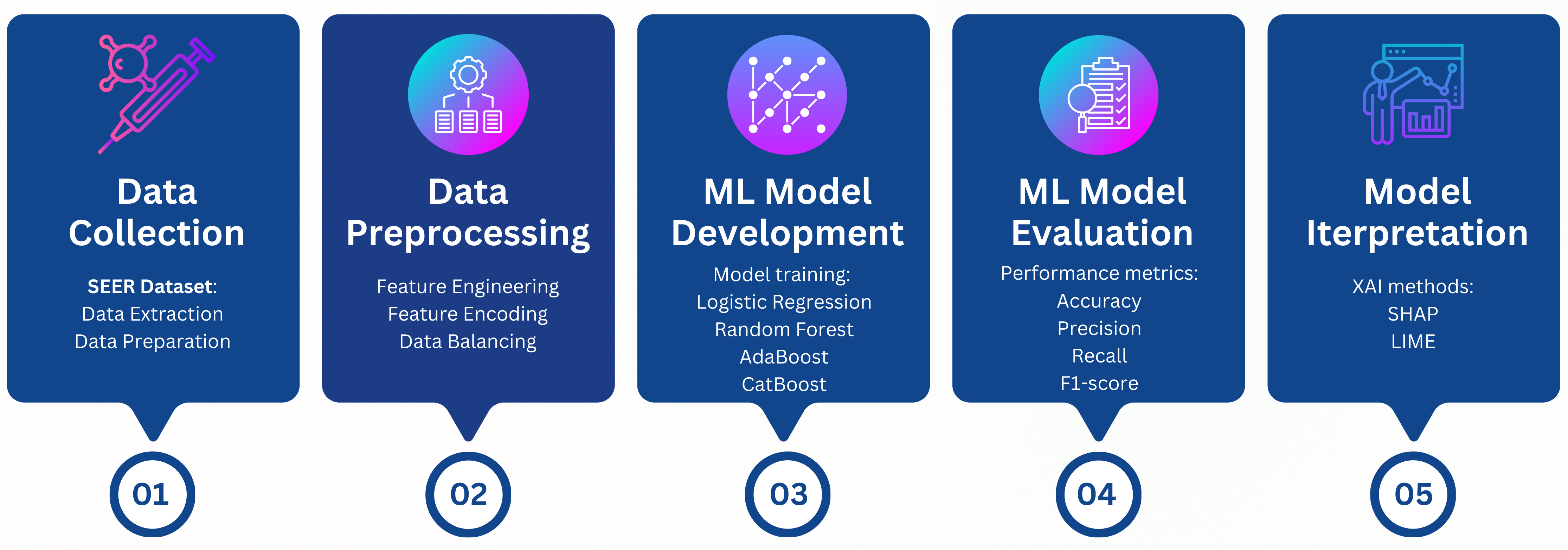}
    \caption{ML workflow, including data collection, preprocessing, model training, evaluation, and interpretation using SHAP~\cite{lundberg2017unified} and LIME~\cite{ribeiro2016should}.}
    \label{fig:pipeline}
\end{figure*}

Responsible ML is essential for advancing the implementation of machine learning in healthcare. Robustness and interpretability have been prioritized in the implementation of responsible machine learning systems~\cite{saraswat2022explainable,rasheed2022explainable,amirian2023explainable,amann2020explainability}. The effective application of computational technology in healthcare is enhanced by mutual awareness of the principles of responsible and explainable AI. The technical and clinical significance of this study can be summarized as follows.

\textit{Technical Significance:} Several ML models are designed for each stage of cancer individually, combined with ML explainability of these models. Thus, the effective features for predicting the survival of patients with specific stages of cancer are better understood. This combination of ML explainability and stepwise prediction helps improve clinical decision-making and patient outcomes.

\textit{Clinical Significance:} Matching ML models with specific stages of cancer increases the effectiveness of clinical prediction. Additionally, the explainability of ML models results in a better understanding of ML-based systems, and helps build trust between doctors and ML. 

\begin{table}[t]
\caption{Overview of the predictor variables used in the machine learning models, including their data types and descriptions.}
\label{tab:features}
\centering
\small
\renewcommand{\arraystretch}{1.1}

\begin{tabular}{
  >{\raggedright\arraybackslash}p{0.40\linewidth}
  >{\raggedright\arraybackslash}p{0.14\linewidth}
  >{\raggedright\arraybackslash}p{0.35\linewidth}
}
\toprule
\textbf{Feature / Variable} & \textbf{Data Type} & \textbf{Description} \\
\midrule

\multicolumn{3}{l}{\textbf{Ordinal Features}} \\
Extension (CS extension code) & Ordinal & Extent of primary tumor spread. \\
Grade & Ordinal & Differentiation level of the tumor cells. \\
Lymph Nodes (CS lymph nodes) & Ordinal & Number of lymph nodes with cancer involvement. \\
Marital Status & Ordinal & Marital status at the time of diagnosis. \\
Radiation & Ordinal & Type of initial radiation therapy received. \\
Surgery Code (RX Summ--Surg Prim Site) & Ordinal & Type of surgery on the primary site. \\
Summary Stage & Ordinal & Overall cancer stage summarizing disease spread. \\
\midrule

\multicolumn{3}{l}{\textbf{Numeric Features}} \\
Age & Numeric & Age at the time of cancer diagnosis. \\
Regional Nodes Examined & Numeric & Number of lymph nodes examined. \\
Regional Nodes Positive & Numeric & Number of lymph nodes positive for cancer. \\
Tumor Size (CS tumor size) & Numeric & Tumor size measured in millimeters. \\
\midrule

\multicolumn{3}{l}{\textbf{Nominal Features}} \\
Behavior Code & Nominal & Indicates tumor aggressiveness. \\
Histologic Type (ICD-O-3) & Nominal & Histological classification of the tumor. \\
Metastasis at Diagnosis (CS mets at dx) & Nominal & Presence of distant metastases at diagnosis. \\
Primary Site & Nominal & Anatomical location of the primary tumor. \\
Race & Nominal & Racial or ethnic background of the patient. \\
Sequence Number & Nominal & Order of this tumor among multiple tumors. \\
Sex (Gender) & Nominal & Gender of the patient as recorded. \\

\bottomrule
\end{tabular}
\end{table}

\section{Materials and Methods}
The ML workflow adopted in this study commenced with the collection of data from the SEER dataset~\cite{seer} for colorectal, stomach, and liver cancers. Following data cleaning, one-hot encoding, and splitting by cancer stage, several machine learning models (Logistic Regression~\cite{hosmer2013applied}, AdaBoost~\cite{freund1996experiments}, Random Forest~\cite{breiman2001random}, and CatBoost~\cite{prokhorenkova2018catboost}) were trained using the selected clinical and demographic features. Hyperparameter tuning was performed using GridSearchCV, and the models were evaluated using the appropriate performance metrics. Finally, to ensure interpretability, SHAP is used for global explanations, whereas LIME  provides local, instance-based insights into the model predictions. The overall process is illustrated in Figure~\ref{fig:pipeline}.


\begin{figure*}[!htb]
\centering
{%
  \includegraphics[width=0.29\textwidth]{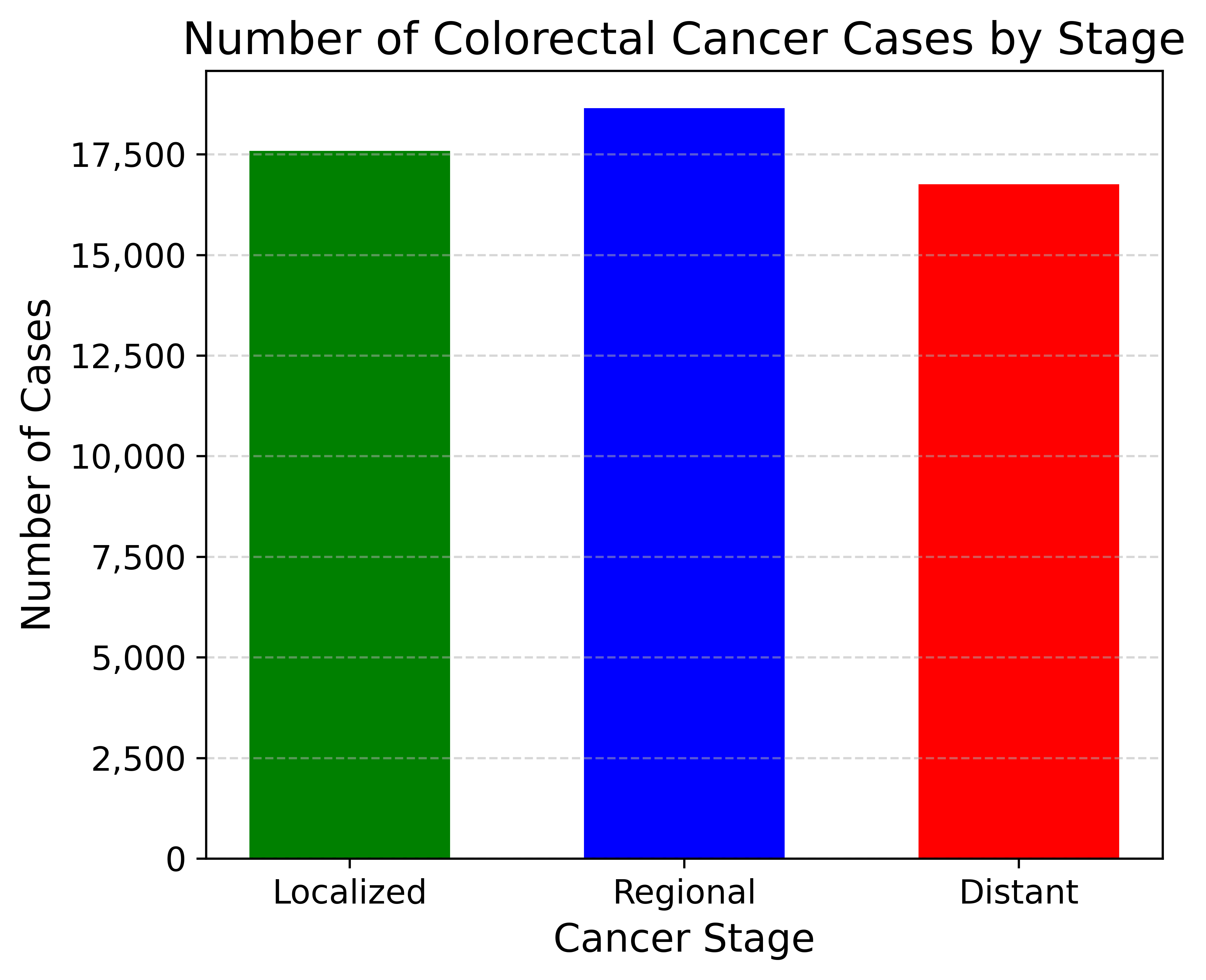}}
\hspace{0.02\textwidth} %
{%
  \includegraphics[width=0.29\textwidth]{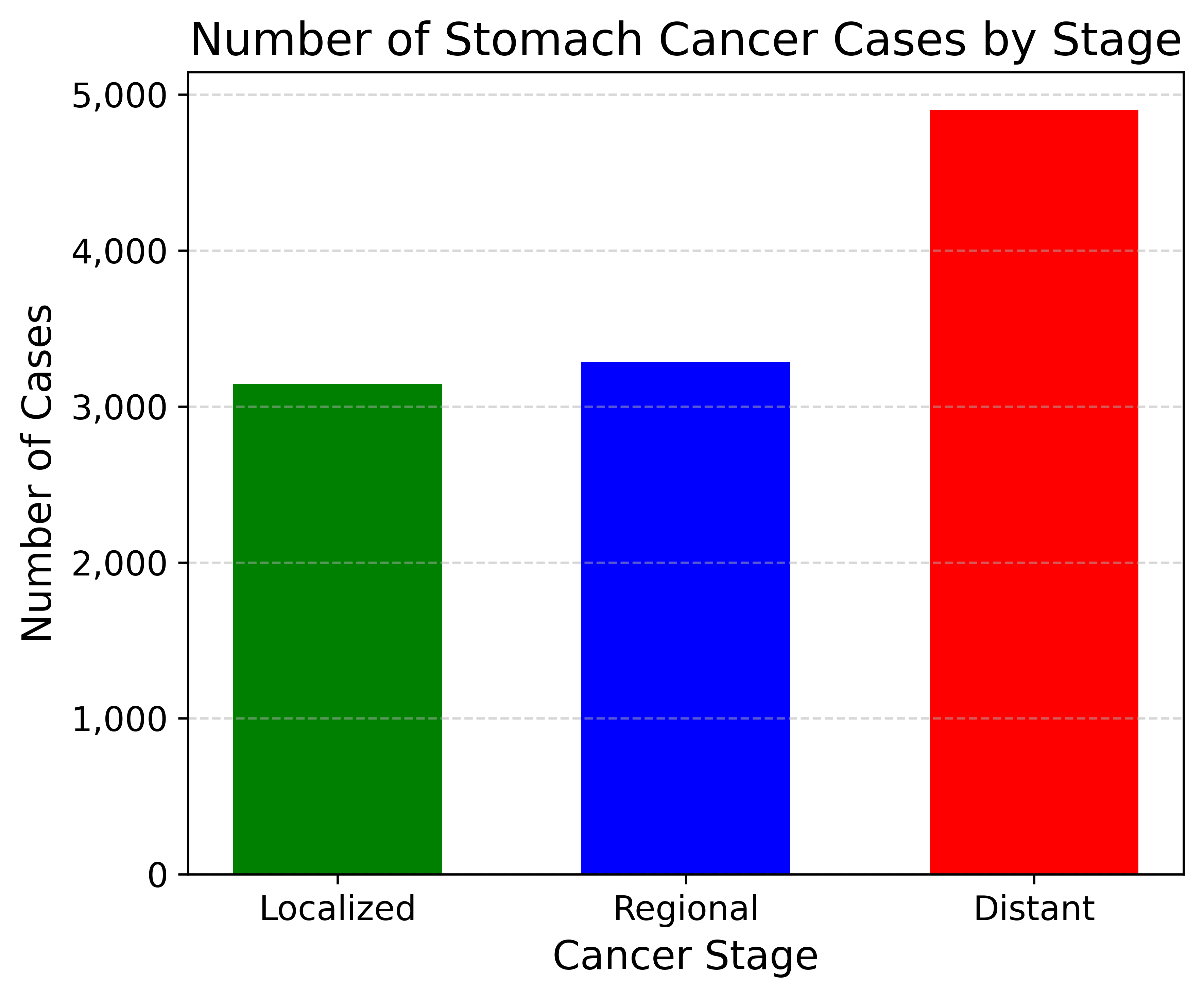}}
\hspace{0.02\textwidth} %
{%
  \includegraphics[width=0.29\textwidth]{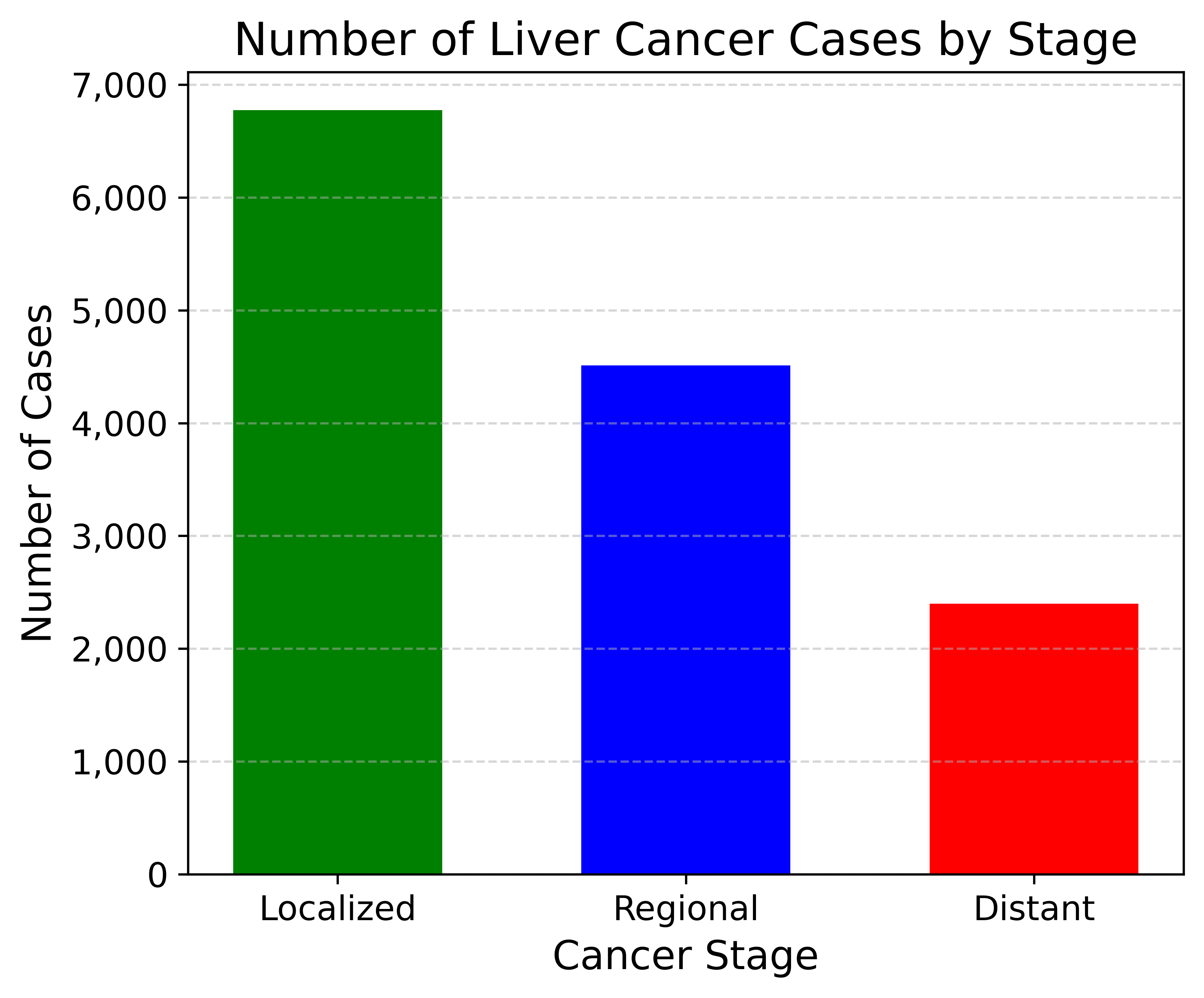}}
\caption{Stage distribution of colorectal, stomach, and liver cancers. The plots illustrate the distribution of cases across stages, emphasizing the differing patterns of disease presentation among cancer types.}
\label{fig:stage_distributions}
\end{figure*}

\subsection{Data source}
This study used data from the Surveillance, Epidemiology, and End Results (SEER) Program~\cite{seer} on gastrointestinal cancers, including colorectal, stomach, and liver cancers, diagnosed between 2004 and 2015 (data accessed March 2025). The SEER program performs annual outlier detection and quality assurance using the SEER Box Plot Outlier Tool (BOT), which identifies and flags unusual values across registries and time~\cite{seerbot}. Rows with missing values in any of the selected features were removed to ensure data quality and consistency across models. Following data preprocessing and stage-based classification, the final datasets were analyzed to determine the distribution of cancer stages for each type.

The features of this dataset included demographic information, tumor characteristics, treatment, and factors associated with patient survival. The variables used to build the ML models are listed in Table~\ref{tab:features}. The features used for modeling were chosen based on previous studies ~\cite{pour2018stage,kate2017stage,kim2013breast} in which they had already been identified and validated, ensuring both clinical relevance and consistency with existing research.

\begin{figure*}[!b]
\centering
{%
  \includegraphics[width=0.31\textwidth]{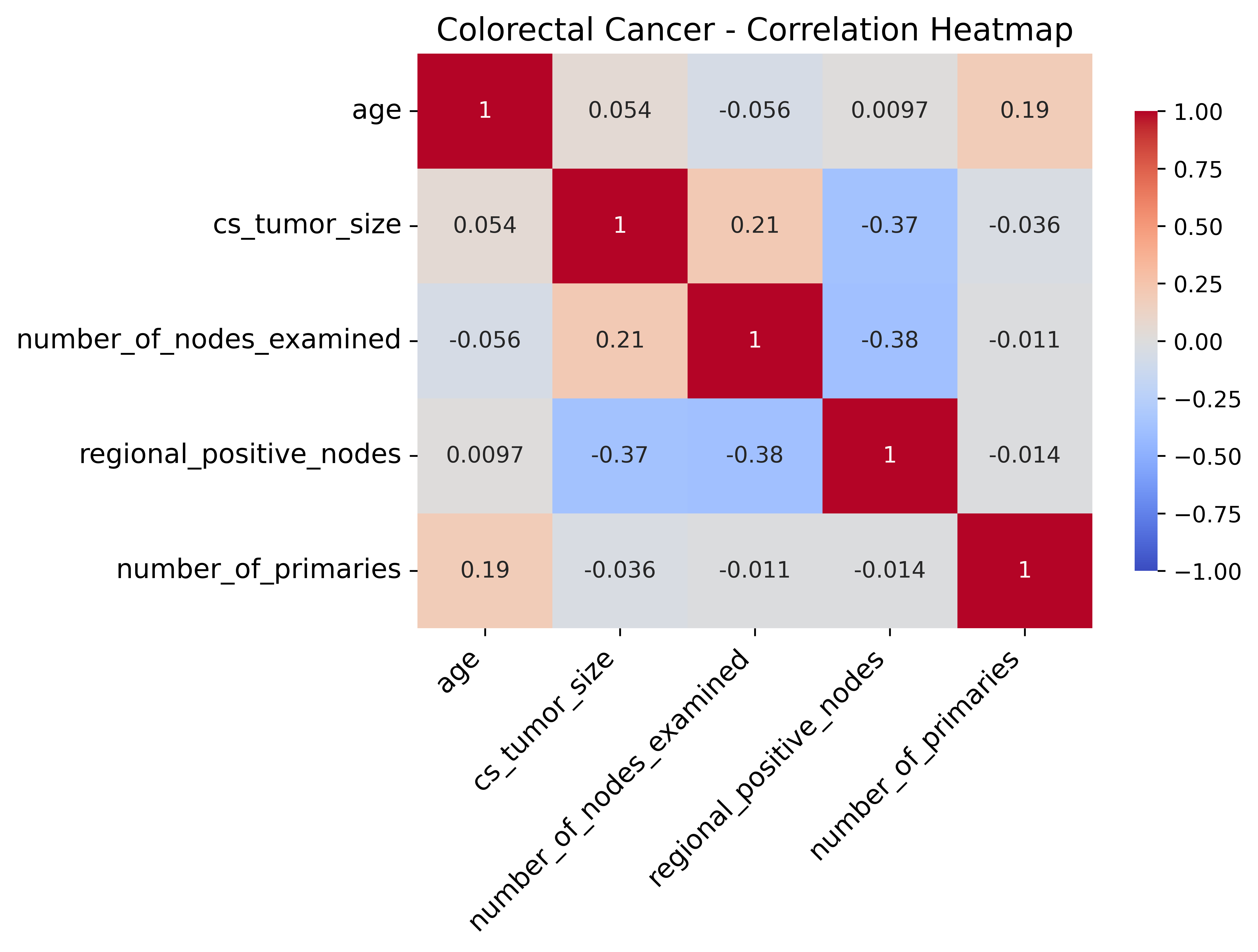}}
{%
  \includegraphics[width=0.31\textwidth]{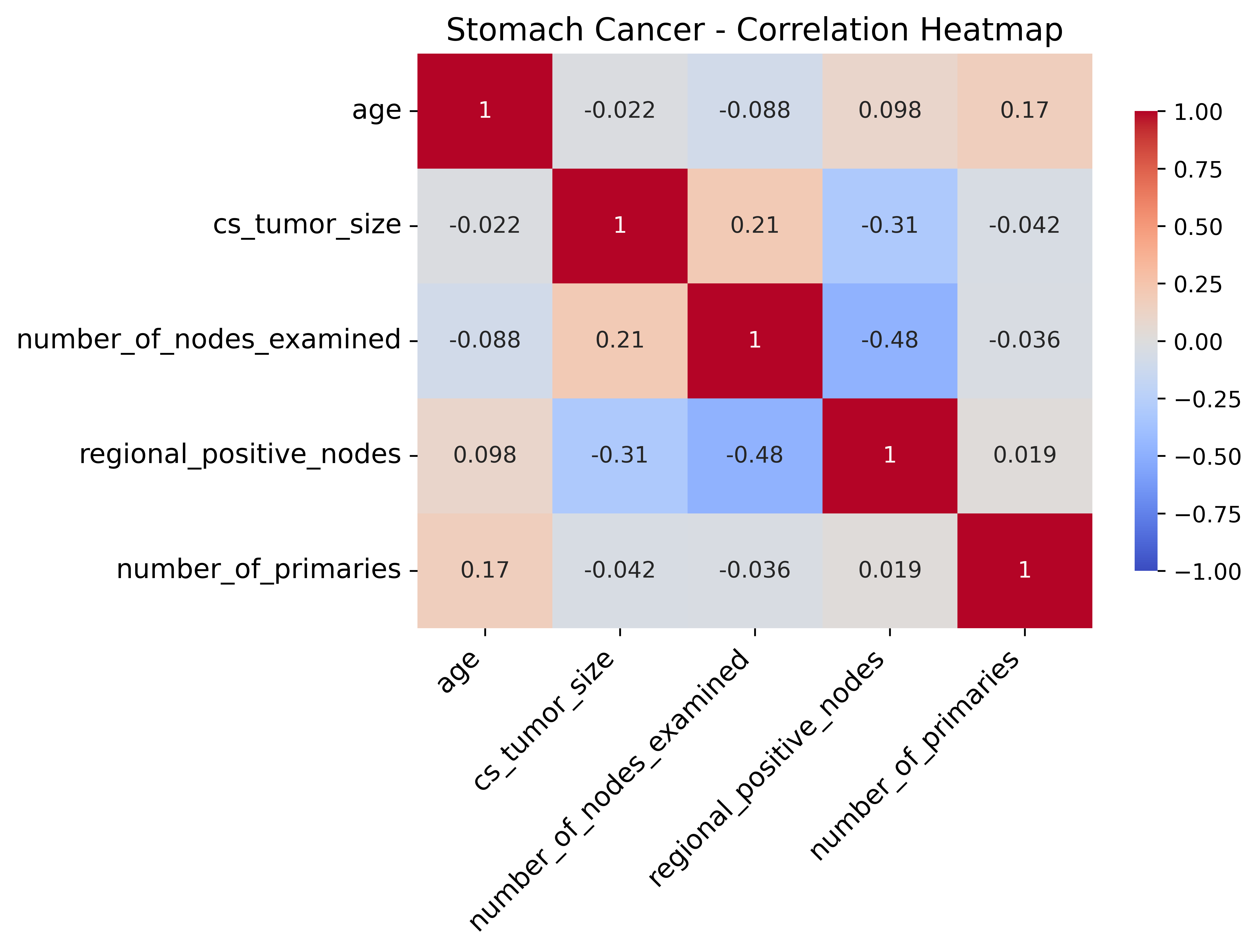}}
\hspace{0.02\textwidth} %
{%
  \includegraphics[width=0.31\textwidth]{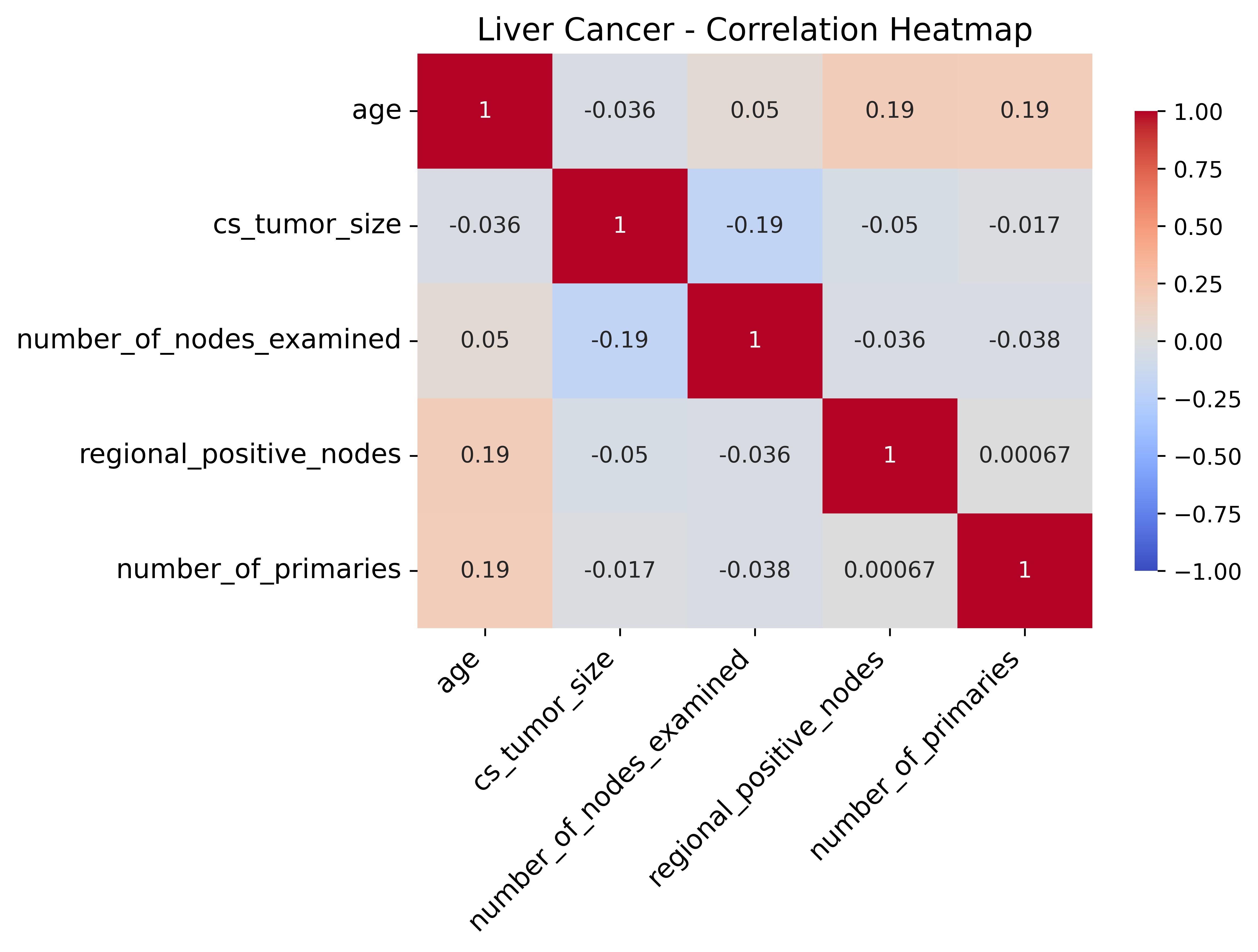}}
\caption{Correlation heatmaps between numerical features for colorectal, stomach, and liver cancers.}
\label{fig:heatmaps}
\end{figure*}

To better understand the relationships between the numerical features, we generated correlation heatmaps for each cancer type (colorectal, stomach, and liver cancer). These heatmaps allow the visualization of pairwise correlations between features, such as age, tumor size, number of examined lymph nodes, number of positive lymph nodes, and number of primaries in the model. This step helped identify potential multicollinearity and provided initial insights into the data structure. The correlation patterns are shown in Figure~\ref{fig:heatmaps}. As the observed correlations were generally weak and no strong multicollinearity was detected, no numerical features were removed from the analysis.

This dataset was prepared through the following processes and was used as input for modeling and analysis in this study.
\subsubsection{Data normalization} To ensure consistent scale across features and improve model convergence, all numerical features were standardized using Z-score normalization~\cite{thakker2024effect}. Specifically, StandardScaler from scikit-learn was used, which transforms each feature to have a mean of 0 and a standard deviation of 1~\cite{thakker2024effect}. This step was integrated into the machine learning pipeline to prevent data leakage during cross-validation.

\subsubsection{Balancing the Dataset} The goal variable, 'Vital status,' exhibited class imbalance. This indicates that one class (e.g., survivors) had  more samples than the other class (non-survivors). To address the class imbalance in the target variable, we adopted class-weighting strategies rather than resampling techniques. This decision aligns with the findings in the literature, where weighting methods are recognized as particularly effective in medical data contexts. According to recent studies~\cite{kalton2003weighting}, class weighting preserves the integrity of the original clinical data and avoids the introduction of artificial patterns, making it a preferred approach in health-related machine learning applications.
    
For models such as Logistic Regression and Random Forest, we used the "balanced" and "balanced\_subsample" options, which assign weights inversely proportional to class frequencies. In the CatBoost model, manual class weights, such as [1, 3] and [1, 5], were tested as hyperparameters using grid search. These settings were evaluated through cross-validation, and the best-performing configuration was selected based on the ROC-AUC scores. This weighting strategy improved prediction performance, particularly for the minority class, while maintaining the integrity of the original dataset.

\subsection{Cancer survivability}
This study determined the gold standard for five-year cancer survivability across cancer types using the rationale in~\cite{pour2018stage}, as shown in Fig.~\ref{fig:five}. This process aimed to establish a reliable standard for measuring the duration of patient survival after diagnosis. To calculate this standard, we used three features: vital status record, survival months, and cause of death, which are available in the SEER dataset~\cite{seer}. Vital status determines whether a patient is alive or dead at a given time. Survival months measure the duration from diagnosis to death or the last follow-up date. The cause of death helps to focus on deaths caused by cancer.

\begin{figure}[H]
\centering
\renewcommand{\arraystretch}{1.1}
\small
\begin{tabular}{|p{0.95\columnwidth}|}
\hline
\rowcolor[gray]{0.6}
\textbf{Five-Year Cancer Survival Standard} \\
\hline
\textbf{if} Survival Months $\geq$ 60 \textbf{and} Vital Status Recode = alive $\rightarrow$ \textbf{survived} \\
\textbf{else if} Survival Months $<$ 60 \textbf{and} Cause of Death = ``Cancer Type'' $\rightarrow$ \textbf{not survived} \\
\textbf{else} patient exclusion \\
\hline
\end{tabular}
\caption{Rule-based criteria for defining five-year cancer survival status~\cite{pour2018stage}.}
\label{fig:five}
\end{figure}

The SEER dataset categorizes cancer into three stages: \textit{localized}, \textit{regional}, and \textit{distant}. In the \textit{localized} stage, cancer remains confined to the organ of origin. In the \textit{regional} stage, it disseminates to adjacent locations, such as lymph nodes or tissues proximal to the organ. The \textit{distant} stage occurs when cancer metastasizes to distant organs or tissues within the body. In this study, we developed models to predict and understand cancer at every stage.

\subsection{Machine learning models construction and evaluation
}
We used four ML models for data analysis: Logistic Regression~\cite{hosmer2013applied}, AdaBoost~\cite{freund1996experiments}, Random Forest~\cite{breiman2001random}, and CatBoost~\cite{prokhorenkova2018catboost}.
A previous study~\cite{pour2018stage} developed stage-specific cancer survivability prediction models using traditional classification algorithms such as Logistic Regression and Decision Trees. Their results demonstrated the effectiveness of these methods in multiple cancer types and clinical stages. We expanded their methodology by incorporating ensemble and boosting-based models, including AdaBoost and CatBoost, to enhance predictive performance and better capture complex patterns in the data.

The ML models were trained and evaluated using stratified K-fold cross-validation, which partitions the dataset while preserving the balance and diversity of the class within each fold. K-fold cross-validation (CV) is the predominant method for assessing the generalization performance of machine learning models and is often preferred over traditional hypothesis testing when the objective is to evaluate predictive reliability rather than statistical significance. This enhancement employs metrics derived directly from machine learning classifications, such as accuracy, which lacks a parametric characterization. The ranges of the hyperparameters explored for each machine learning model during the grid search process are summarized in Table~\ref{tab:hyperparameters}. Our hyperparameter selection was inspired by~\cite{kamble2025predicting}, with modifications and additions, including the CatBoost model and extended parameter ranges for some of the classifiers.

\begin{table}[!htb]
\caption{Hyperparameter ranges used during grid search for each ML model.}
\label{tab:hyperparameters}
\centering
\setlength{\tabcolsep}{4pt} %
\renewcommand{\arraystretch}{1.2} %
\resizebox{0.490\textwidth}{!}{%
\begin{tabular}{p{0.15\textwidth} p{0.15\textwidth} p{0.20\textwidth}}
\toprule
\textbf{Model} & \textbf{Parameter} & \textbf{Values Tested} \\
\midrule

Logistic Regression & C & 0.001, 0.01, 0.1, 1, 10 \\
                    & Class Weight & None, balanced \\[5pt]

AdaBoost            & n\_estimators & 50, 100, 200 \\
                    & Learning Rate & 0.01, 0.1, 1.0 \\
                    & Algorithm & SAMME, SAMME.R \\[5pt]

Random Forest       & n\_estimators & 100, 200 \\
                    & Max Depth & 3, 5, 7 \\
                    & Min Samples Split & 2, 5 \\
                    & Min Samples Leaf & 1, 2, 4 \\
                    & Class Weight & balanced, balanced\_subsample \\[5pt]

CatBoost            & Iterations & 100, 200 \\
                    & Depth & 3, 5, 7 \\
                    & Learning Rate & 0.03, 0.1 \\
                    & L2 Leaf Reg & 1, 3, 5 \\
                    & Class Weights & [1,1], [1,3], [1,5] \\
\bottomrule
\end{tabular}}
\end{table}





\subsection{Models explainability}
To address the comprehensibility and interpretability of ML models for cancer survival prediction, we employed two prevalent techniques in explainable artificial intelligence (XAI): Shapley Additive Explanations (SHAP)~\cite{lundberg2017unified} and Local Interpretable Model-agnostic Explanations (LIME)~\cite{ribeiro2016should}. 

SHAP provides global interpretability by quantifying the contribution of each feature (e.g., tumor size and cancer stage) to predictions across the dataset. Complementing this global view, LIME was applied for local interpretability to the individual patients with the lowest predicted survival probability, evaluated across all stages combined and within each clinical stage (localized, regional, distant) for all studied cancer types by locally approximating the model with a simple, interpretable surrogate.

The integration of SHAP and LIME in this study provides a comprehensive and interpretable approach. SHAP offers broad, feature-focused explanations, whereas LIME provides detailed case-specific insights. This combined method improves the model's transparency, boosting trust and understanding when predicting cancer survival rates.

Although XAI improves model transparency and provides valuable insights, it also has limitations. As noted by Salih et al.~\cite{salih2025perspective}, the application of methods such as SHAP and LIME in sensitive fields, requires caution. Misinterpretation of their outputs or reliance on them without a clear understanding of their underlying assumptions can result in serious errors and unintended consequences for patients.

\section*{Acknowledgment}
The authors thank the SEER program for providing access to the datasets. The authors gratefully acknowledge Prof. Ahmad P. Tafti, PhD, FAMIA, at the University of Pittsburgh, for his support and comments on this research.

\section{Results}
The experimental setup involved stratifying the SEER dataset~\cite{seer} into localized, regional, and distant stages for colorectal, stomach, and liver cancer. The machine learning models were trained using stratified K-fold cross-validation and evaluated based on metrics such as accuracy, precision, recall, F1-score, and AUC-ROC. CatBoost exhibited the best performance across most stages and cancer types.

\subsection{Data Preprocessing}
The breakdown of cases in the local, regional, and distant stages of colorectal, liver, and stomach cancers is summarized below. 

\textit{Colorectal Cancer (n = 53,100)}: The colorectal cancer dataset comprised a relatively balanced distribution across three main stages. Localized tumors accounted for 33.1\% of cases (n = 17,582), followed closely by regional stage with 35.6\% (n = 18,884). Distant-stage cases accounted for 31.3\% (n = 16,634), reflecting a broad representation of disease progression.

\textit{Stomach Cancer (n = 11,350)} : The stomach cancer dataset displayed a distinct distribution, with distant-stage cases being the most prevalent at 43.2\% (n = 4,903). Regional cases comprised 29.0\% (n = 3,292), whereas localized cancers comprised 27.8\% (n = 3,155), indicating a tendency toward late-stage diagnoses in this population.

\textit{Liver Cancer (n = 13,700)}: In the liver cancer cohort, localized cases were the most common, representing 49.6\% of the total (n = 6,791), followed by regional cases at 33.0\% (n = 4,520). Distant-stage cancers were less frequent, accounting for 17.4\% (n = 2,389), suggesting a higher proportion of early detection in this group.

The stage-wise distributions of these cancer types are shown in Figure~\ref{fig:stage_distributions}.

\begin{table*}[!htb]
\caption{Performance of ML models for survival prediction across cancer types and stages.}
\label{tab:cancer_model_metrics_compact}
\centering
\renewcommand{\arraystretch}{1.2}
\small
\resizebox{0.99\textwidth}{!}{
\begin{tabular}{ll
ccccc 
ccccc 
ccccc 
ccccc 
}
\toprule
\multirow{2}{*}{\textbf{Cancer}} & \multirow{2}{*}{\textbf{Stage}} 
& \multicolumn{5}{c}{\textbf{Logistic Regression}} 
& \multicolumn{5}{c}{\textbf{Random Forest}} 
& \multicolumn{5}{c}{\textbf{AdaBoost}} 
& \multicolumn{5}{c}{\textbf{CatBoost}} \\
\cmidrule(lr){3-7} \cmidrule(lr){8-12} \cmidrule(lr){13-17} \cmidrule(lr){18-22}
& & Acc & Prec & Rec & F1 & AUC & Acc & Prec & Rec & F1 & AUC & Acc & Prec & Rec & F1 & AUC & Acc & Prec & Rec & F1 & AUC \\
\midrule

\multirow{3}{*}{Colorectal}
 & Localized  & 0.858 & 0.862 & 0.987 & 0.920 & 0.805  & 0.859 & 0.865 & 0.982 & 0.920 & 0.817  & 0.857 & 0.870 & 0.972 & 0.918 & 0.813  & 0.864 & 0.873 & 0.977 & 0.922 & 0.821 \\
 & Regional   & 0.677 & 0.642 & 0.908 & 0.752 & 0.758  & 0.700 & 0.658 & 0.924 & 0.769 & 0.791  & 0.710 & 0.726 & 0.774 & 0.749 & 0.797  & 0.722 & 0.682 & 0.912 & 0.780 & 0.809 \\
 & Distant    & 0.886 & 0.323 & 0.482 & 0.386 & 0.821  & 0.913 & 0.416 & 0.406 & 0.411 & 0.839  & 0.928 & 0.590 & 0.092 & 0.160 & 0.845  & 0.912 & 0.422 & 0.486 & 0.451 & 0.860 \\
\midrule

\multirow{3}{*}{Stomach}
 & Localized  & 0.887 & 0.893 & 0.945 & 0.918 & 0.937  & 0.901 & 0.919 & 0.936 & 0.927 & 0.951  & 0.887 & 0.907 & 0.927 & 0.917 & 0.952  & 0.895 & 0.880 & 0.976 & 0.926 & 0.956 \\
 & Regional   & 0.758 & 0.527 & 0.733 & 0.613 & 0.824  & 0.822 & 0.632 & 0.767 & 0.693 & 0.865  & 0.830 & 0.731 & 0.552 & 0.629 & 0.877  & 0.830 & 0.650 & 0.756 & 0.699 & 0.880 \\
 & Distant    & 0.976 & 0.741 & 0.541 & 0.625 & 0.953  & 0.963 & 0.512 & 0.595 & 0.550 & 0.942  & 0.962 & 0.500 & 0.676 & 0.575 & 0.938  & 0.968 & 0.558 & 0.784 & 0.652 & 0.955 \\
\midrule

\multirow{3}{*}{Liver}
 & Localized  & 0.751 & 0.582 & 0.816 & 0.679 & 0.842  & 0.772 & 0.614 & 0.794 & 0.693 & 0.854  & 0.785 & 0.641 & 0.761 & 0.696 & 0.860  & 0.781 & 0.634 & 0.768 & 0.694 & 0.868 \\
 & Regional   & 0.905 & 0.524 & 0.530 & 0.527 & 0.826  & 0.920 & 0.635 & 0.482 & 0.548 & 0.846  & 0.899 & 0.495 & 0.554 & 0.523 & 0.839  & 0.919 & 0.614 & 0.518 & 0.562 & 0.860 \\
 & Distant    & 0.977 & 0.500 & 0.545 & 0.522 & 0.899  & 0.975 & 0.462 & 0.545 & 0.500 & 0.890  & 0.981 & 0.600 & 0.545 & 0.571 & 0.880  & 0.987 & 0.857 & 0.545 & 0.667 & 0.811 \\
\bottomrule
\end{tabular}}
\end{table*}
    
\begin{figure*}[htb!]
    \centering
    \begin{subfigure}[b]{0.32\textwidth}
        \centering
        \includegraphics[width=\textwidth]{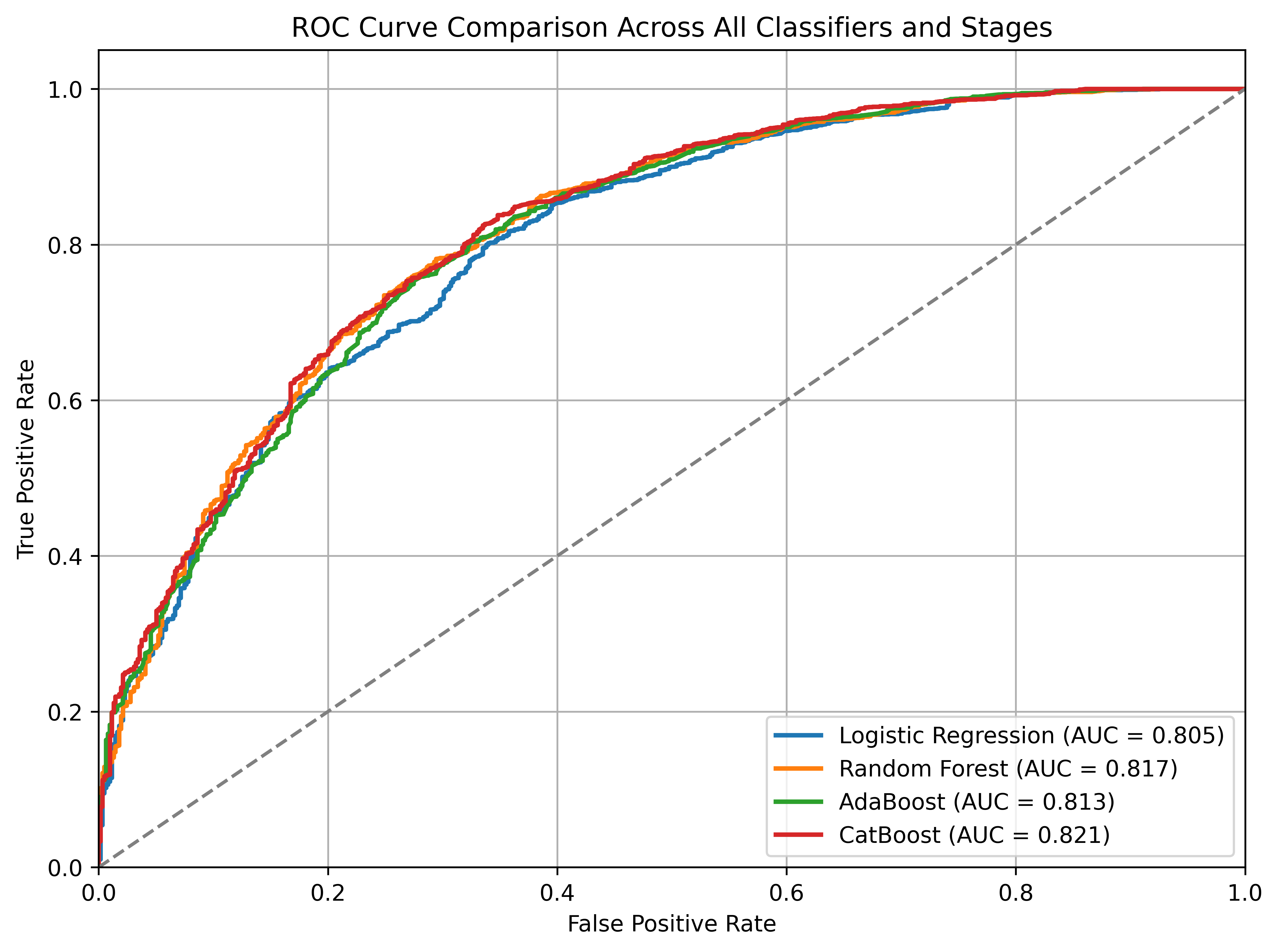}
        \caption{Localized stage}
        \label{fig:ROC_curves_sub1}
    \end{subfigure}
    \hfill
    \begin{subfigure}[b]{0.32\textwidth}
        \centering
        \includegraphics[width=\textwidth]{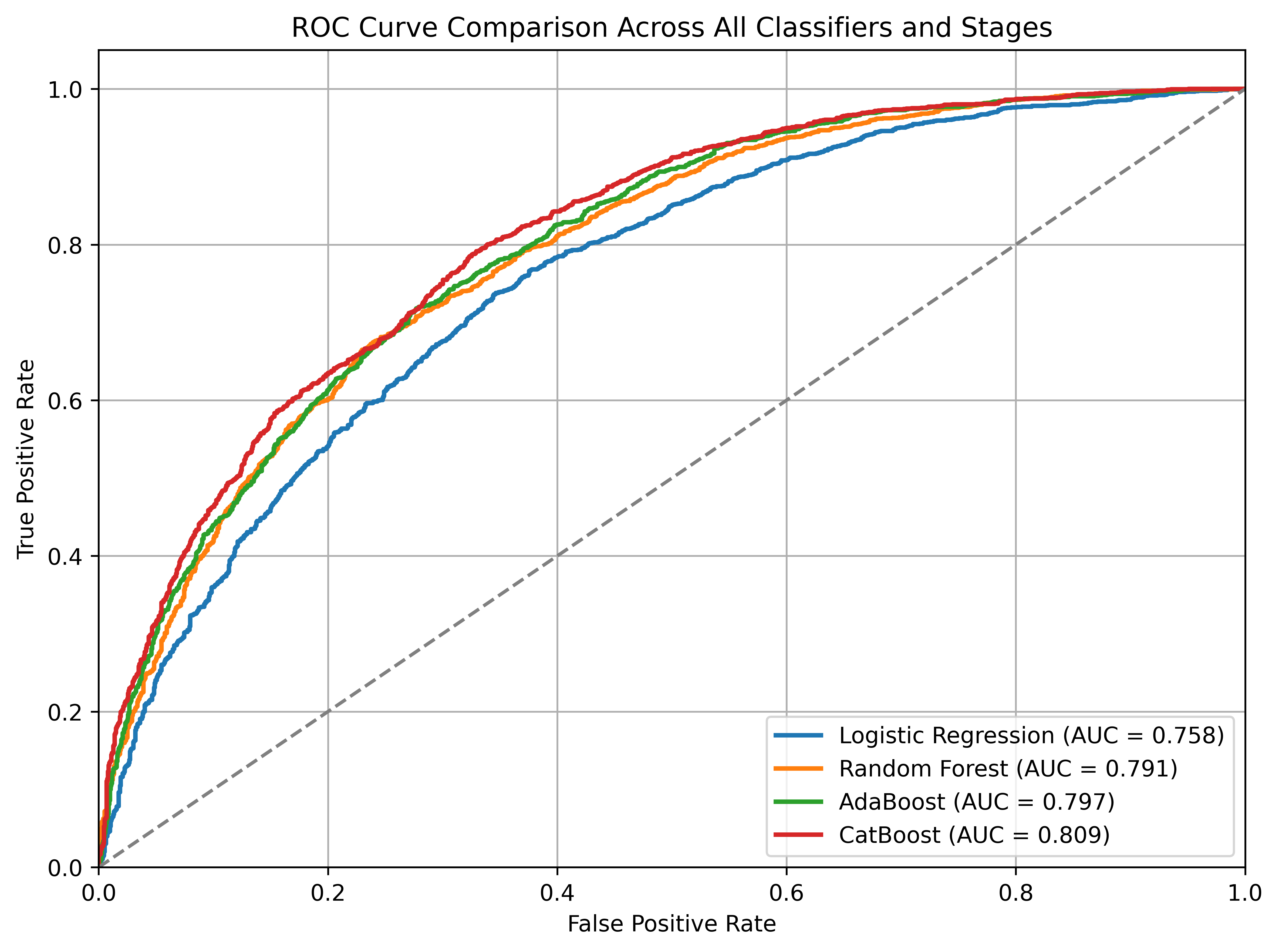}
        \caption{Regional stage}
        \label{fig:ROC_curves_sub2}
    \end{subfigure}
    \hfill
    \begin{subfigure}[b]{0.32\textwidth}
        \centering
        \includegraphics[width=\textwidth]{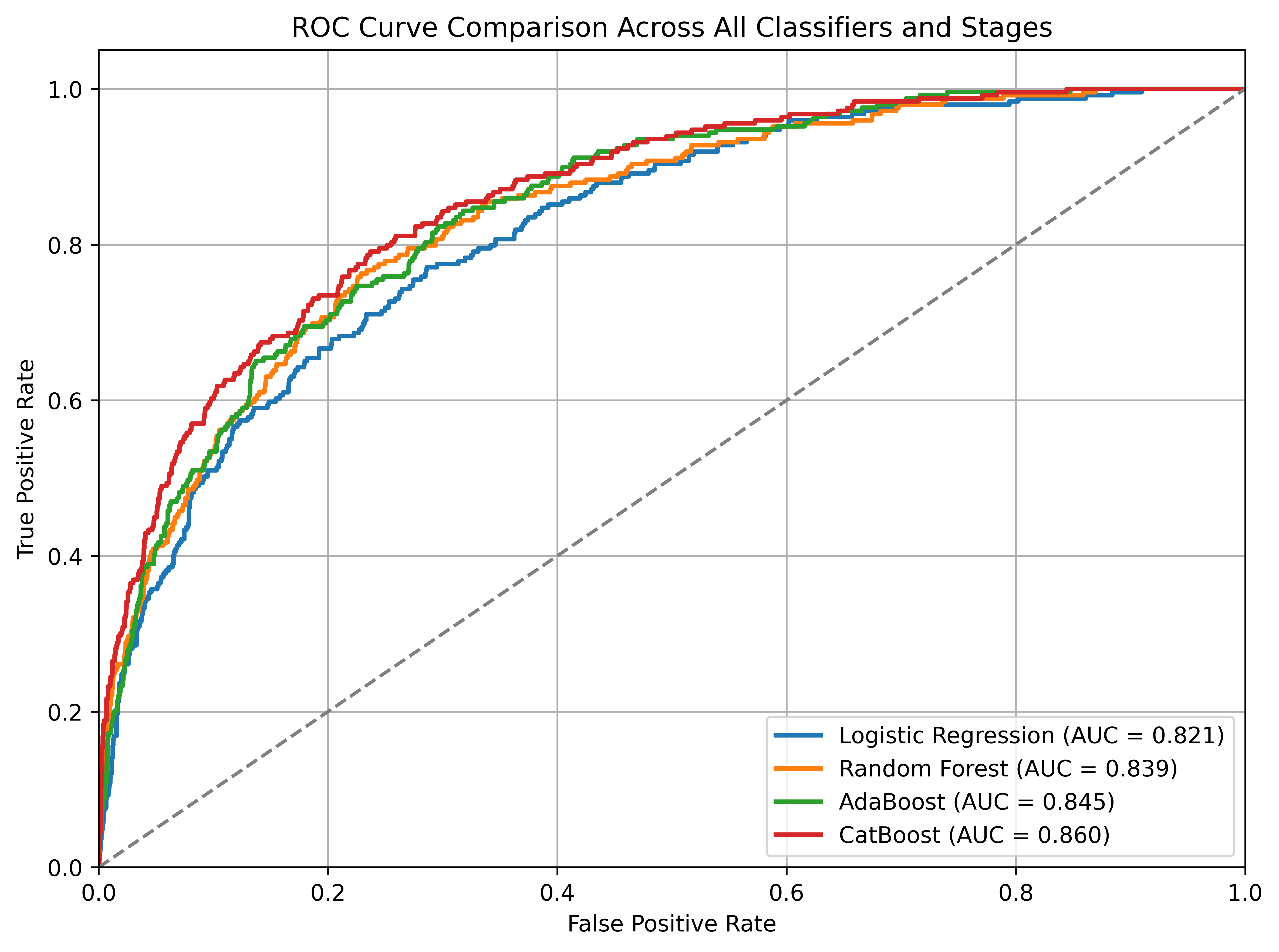}
        \caption{Distant stage}
        \label{fig:ROC_curves_sub3}
    \end{subfigure}

    \caption{The receiver operating characteristic (ROC) curves of the four models (Logistic Regression, Random Forest, AdaBoost, CatBoost) for colorectal cancer across different stages: (a) Localized, (b) Regional, and (c) Distant.}
    \label{fig:ROC_curves}
\end{figure*}

\subsection{Models performance}
In this section, the performance of machine learning models in predicting the survival of patients with colorectal, stomach, and liver cancer is discussed. Three distinct stage categories, namely Localized, Regional, and Distant, were used to analyze each type of cancer. The models were evaluated based on five standard metrics: Accuracy, Precision, Recall, F1-score, and AUC-ROC. We adopted the area under the Receiver Operating Characteristic curve (AUC) as our primary evaluation metric. The ROC curve plots the relationship between the true positive rate (i.e., the proportion of correctly identified positive cases, also referred to as sensitivity) and the false-positive rate (i.e., the proportion of negative cases incorrectly classified as positive, equal to 1-specificity). In our setup, the positive class corresponds to “survived,” while the negative class denotes “not survived.” Because the ROC curve is unaffected by class distribution [22], it provides a more reliable performance measure for our task than metrics such as precision or recall. To condense the results from the large number of evaluations, we report only the AUC value, which captures the overall performance of the ROC curve. An ideal classifier achieves an AUC of 1. Table~\ref{tab:cancer_model_metrics_compact} illustrates the results. CatBoost achieved the best performance among all evaluated models, yielding the highest AUC-ROC scores with balanced results across all cancer types. Given its superior predictive performance, CatBoost was further examined for its explainability.

In predicting survival for \textit{colorectal cancer}, the CatBoost model achieved the best overall performance and showed more balanced results across different stages, particularly in the regional stage, compared to Logistic Regression and AdaBoost. Across all three stages, ensemble-based models consistently outperformed Logistic Regression in the ROC analysis. Figure~\ref{fig:ROC_curves} presents the ROC curves for~\textit{colorectal cancer} across the three stages (localized, regional, and distant). As shown in Fig.~\ref{fig:ROC_curves_sub1}, CatBoost achieved the highest AUC (0.821), closely followed by Random Forest (0.817) and AdaBoost (0.813). In the localized stage (Fig.\ref{fig:ROC_curves_sub1}), CatBoost achieved the highest AUC (0.821), closely followed by Random Forest (0.817) and AdaBoost (0.813). In the regional stage (Fig.\ref{fig:ROC_curves_sub2}), a slight decline was observed, with CatBoost still leading (0.809), while Logistic Regression dropped to 0.758. In the distant stage (Fig.~\ref{fig:ROC_curves_sub3}), performance improved for all models, with CatBoost again achieving the best result (AUC-ROC=0.860). These results highlight the robustness and superior predictive power of ensemble methods, particularly that of CatBoost. In the distant stage, CatBoost also remained more stable (AUC-ROC=0.860). 

For \textit{stomach cancer}, CatBoost and AdaBoost delivered the best overall results (AUC-ROC $\approx$ 0.96). In the localized stage, Random Forest achieved the highest accuracy and F1-score, while CatBoost recorded the highest AUC-ROC and recall. In the regional stage, Logistic Regression performed the weakest, whereas CatBoost provided more balanced outcomes. In the distant stage, CatBoost again proved to be the most effective, with Recall=0.784 and F1=0.652.

For~\textit{liver cancer}, CatBoost demonstrated a consistently strong performance across stages. While Logistic Regression achieved higher recall in some cases, its low precision limited its effectiveness. CatBoost and AdaBoost provided more balanced results overall, and in the distant stage, CatBoost was the most effective, achieving the highest precision (0.857) and an F1 score of 0.667.

Overall, across all three cancer types and most stages, CatBoost consistently exhibited the most stable and well-balanced performance in survival prediction.



\begin{figure*}[!htb]
\centering
\subfloat[SHAP  beeswarm sumary plot for the prediction of survivability of colorectal cancer at localized stage using CatBoost. Advanced age and greater extension drive the prediction toward \textit{Dead}, with small countervailing effects from selected histologic codes.\label{fig:case_shap_cr_loc}]{%
  \includegraphics[width=0.40\textwidth]{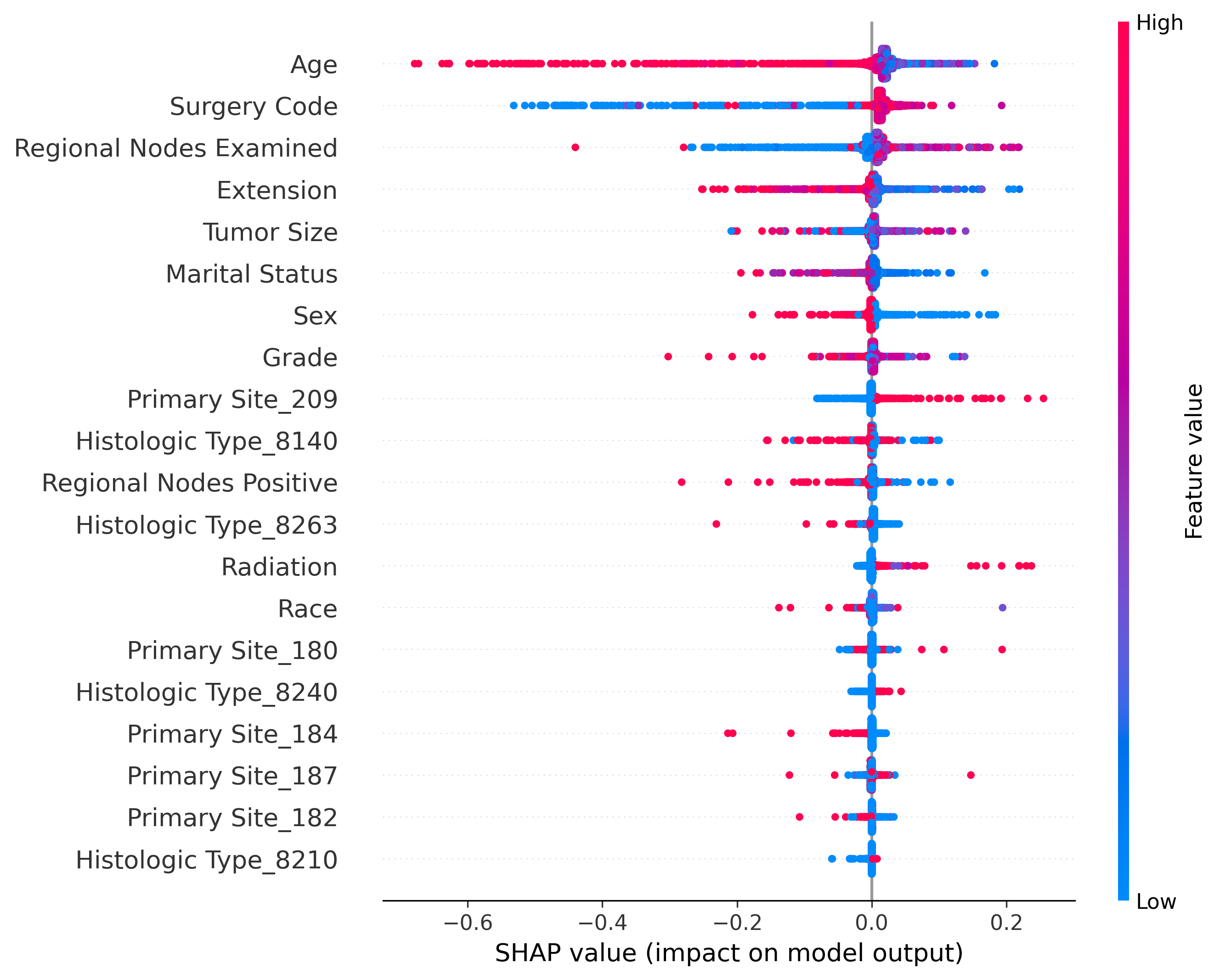}}
\hspace{0.02\textwidth} %
\subfloat[LIME explainability for a single instance (patient). \textit{Regional Nodes Examined} reported as zero, together with age and extension, are the strongest local drivers toward \textit{Dead}.\label{fig:case_lime_cr_loc}]{%
  \includegraphics[width=0.55\textwidth]{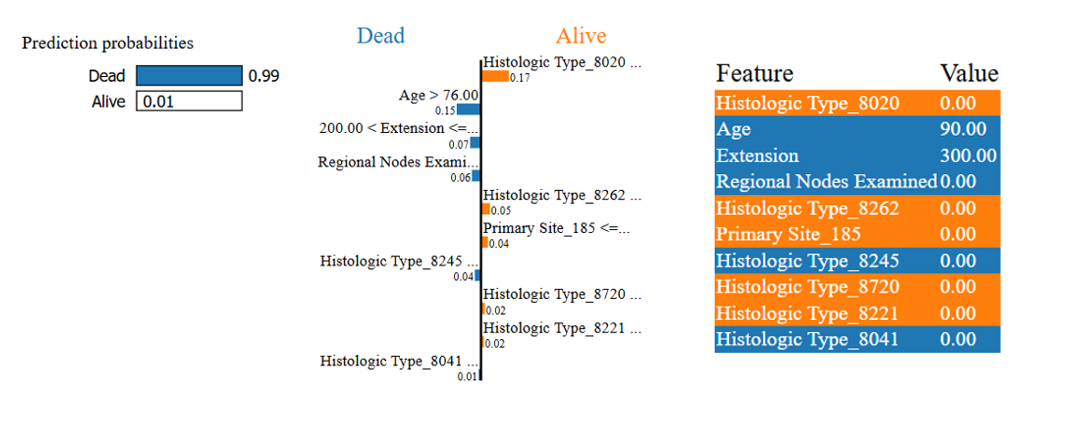}}
\caption{Case-level explanations for a representative patient with \textit{colorectal} cancer at the \textit{localized} stage : (a) SHAP summarizes global impact directions, whereas (b) LIME provides a local, instance-specific rationale. 
}
\label{fig:case_cr_loc_pair}
\end{figure*}

\subsection{Explainability analysis and visualization}
\begin{figure}[!htb]
    \centering
    \includegraphics[width=0.65\textwidth,keepaspectratio]{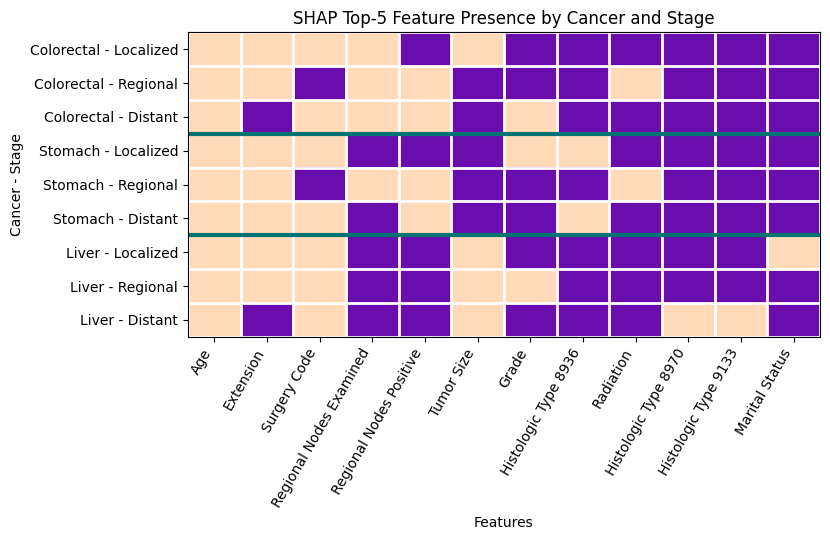}
    \caption{Binary heatmap of top five SHAP features across cancer types (Colorectal, Stomach, and Liver) and stages (Localized, Regional, and Distant). Orange cells indicate that the feature was among the top five most important features for the corresponding cancer-stage group.}
    \label{fig:SHAP5top}
\end{figure}

As CatBoost consistently demonstrated superior predictive performance across all three cancer types and stages, explainability analyses were conducted on this model. Figure~\ref{fig:SHAP5top} shows the top five SHAP-ranked predictors for each cancer type and clinical stage. In localized disease, initial treatment (Surgery Code) and tumor burden (Tumor Size and Extension), together with age, dominate. In Regional disease, Age and Extension consistently appeared among the top features across all three cancers, whereas lymph node indicators, namely Regional Nodes Positive and Regional Nodes Examined, were prominent mainly in colorectal and stomach cancers. Radiation also appeared as one of the top features for these two cancers. In Distant disease, histologic subtypes (SEER codes 8936 for stomach; 9133 and 8970 for liver) emerged beyond Age and Surgery Code, while nodal variables remained important in colorectal cancer.

\begin{figure*}[!htb]
    \centering
    \includegraphics[width=0.99\textwidth,keepaspectratio]{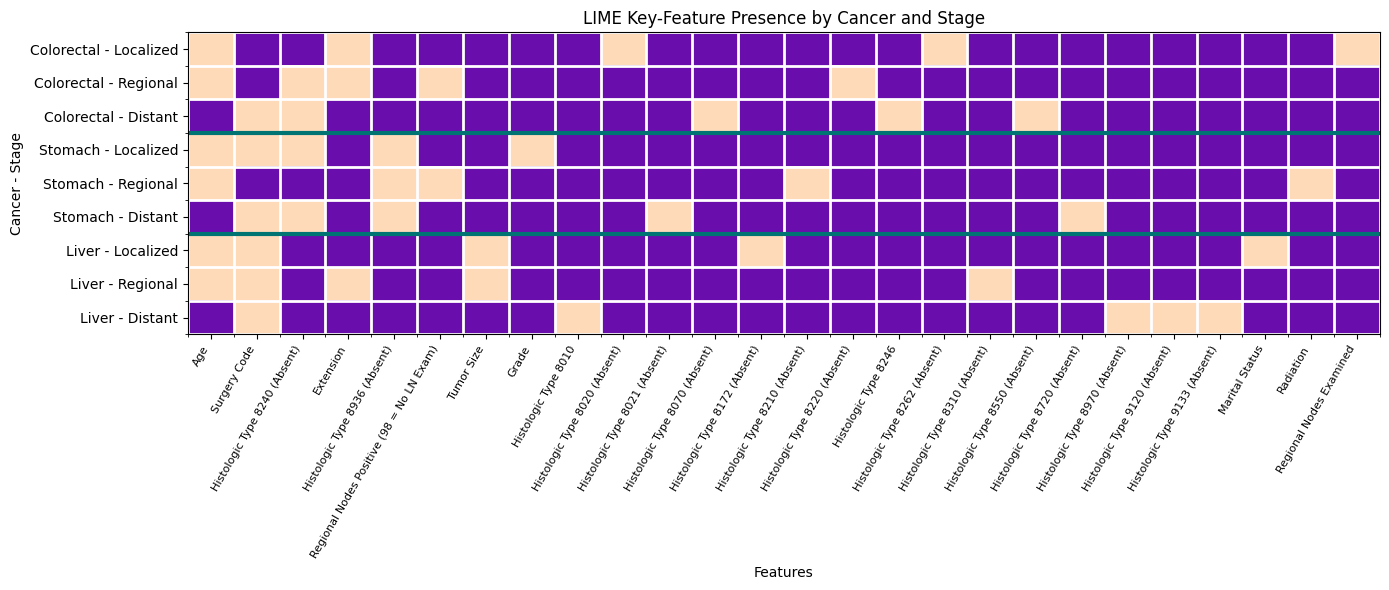}
    \caption{Binary heatmap showing the presence of top-5 LIME features across cancer types (Colorectal, Stomach, and Liver) and stages (Localized, Regional, and Distant). Orange cells indicate that the feature was among the top-5 most important features for the corresponding cancer-stage group.}
    \label{fig:Lime5top}
\end{figure*}

Figure~\ref{fig:Lime5top} shows the presence of LIME-based key features across colorectal, stomach, and liver cancers stratified by disease stage. The orange cells indicate that a given feature was among the top local predictors in the cancerstage group.

Several general patterns were observed. Age consistently appeared across nearly all cancers and stages as an influential factor, reaffirming its well-established role as a strong prognostic indicator of cancer. The surgical code, in particular its absence (value = 0), was prominently highlighted in stomach and liver cancers at multiple stages, underscoring the clinical importance of surgical treatment in the prediction of survival.

Among the lymph node–related variables, the category Regional Nodes Positive with the value of ninety-eight, which in the SEER dataset denotes that no pathological examination of lymph nodes was performed, emerged exclusively in the Regional stage for colorectal and stomach cancers and was absent in other stages. This pattern is clinically logical, as nodal status is a key determinant of regional disease, and the absence of evaluation (code 98) may reflect diagnostic or therapeutic limitations associated with poorer outcomes.

Another notable group of variables consisted of histologic subtype codes (e.g., 8240 and 8936 for stomach; 9133 and 8970 for liver), which appeared predominantly in metastatic (distant-stage) disease cases. Their prominence suggests that tumor biology and subtype heterogeneity exert a greater influence on survival in advanced stages.

Among the indirect clinical factors, marital status appeared only once in the localized stage of liver cancer. This limited occurrence indicates that marital status does not play a primary role in survival prediction but may be relevant in specific stage-dependent contexts.

To illustrate patient-level explanations, we present a representative localized colorectal cancer case (Figs. \ref{fig:case_shap_cr_loc}–\ref{fig:case_lime_cr_loc}). The patient was 90 years old with an extension value of approximately 300 and without any lymph node evaluation, with Regional Nodes Examined reported as zero. The SHAP plot indicates that advanced age and greater local extension are the dominant drivers toward “Dead,” whereas several histologic codes provide small countervailing contributions toward “Alive.” In these SHAP visualizations, red and blue points represent high and low feature values, respectively, clarifying how a patient’s advanced age and extensive tumor involvement strongly push the prediction toward death. The LIME explanation for the same case corroborates this logic at the local level, where Age, Extension, and the absence of nodal evaluation are the main contributors pushing the prediction toward “Dead,” with modest positive influence from a few histologic subtypes, such as 8020 and 8262. This case exemplifies how global (SHAP) and local (LIME) explanations align with cohort-level findings while transparently reconstructing the model’s reasoning at the individual level, thereby supporting trust for clinical use.

\section{Discussion}
Table~\ref{tab:mean_compact2_pval} demonstrates the statistically significant differences in the mean values of key clinical features between survivors and nonsurvivors across all cancer types studied. These findings are consistent with the model interpretation results obtained using SHAP and LIME. Non-survivors were generally older, had more positive lymph nodes, and presented with larger tumors than those in survivors. Conversely, survivors had a higher number of examined lymph nodes, suggesting that better diagnostic and therapeutic procedures were performed. This concordance between statistical trends and model interpretability enhances confidence in the clinical relevance of machine learning model predictions.

\begin{table}[h]
\centering
\caption{Comparison of mean values between survivors and non-survivors with p-values.}
\label{tab:mean_compact2_pval}
\resizebox{0.70\textwidth}{!}{%
\begin{tabular}{lcccc|c}
\toprule
\textbf{Cancer Type} & \textbf{Age} & \textbf{Nodes Exam.} & \textbf{Nodes Pos.} & \textbf{Tumor Size} & \textbf{p-Value} \\
\midrule
\multicolumn{6}{c}{\textbf{Survivors}} \\
Colorectal & 63.2 & 16.0 & 18.9 & 32.5 & 0.0 \\
Stomach    & 60.5 & 11.1 & 43.3 & 27.9 & 0.0 \\
Liver      & 55.6 & 2.5  & 84.1 & 36.2 & 0.0 \\
\midrule
\multicolumn{6}{c}{\textbf{Non-survivors}} \\
Colorectal & 70.0 & 13.8 & 37.8 & 38.9 & 0.0 \\
Stomach    & 68.5 & 7.1  & 70.0 & 31.1 & 0.0 \\
Liver      & 64.9 & 6.8  & 96.3 & 47.8 & 0.0 \\
\bottomrule
\end{tabular}%
}
\end{table}

\section{Conclusion}
In this study, survival prediction for patients with colorectal, stomach, and liver cancers was carried out using stage-specific modeling (Localized, Regional, and Distant). The results showed that model performance varied across stages, and stage-wise analysis better reflected the heterogeneity of patient survival. For example, in colorectal cancer, although the models performed well in the early stages, their sensitivity dropped noticeably in the advanced stage. Similar patterns were observed in stomach and liver cancers, where differences between early and advanced stages became clear only through stage-specific modeling. These findings indicate that stage-based analysis prevents overly optimistic estimates and provides a more accurate and clinically meaningful picture of survival patterns.

Beyond predictive performance, this study emphasizes model explainability. Using SHAP and LIME, we showed that the models relied primarily on well-established prognostic features, such as age, surgery, tumor size, extension, and lymph node involvement. These findings align with prior evidence, where older age has consistently been associated with increased mortality risk~\cite{lin2023impact}, while surgical treatment and a higher number of lymph nodes examined have been linked to improved survival~\cite{kim2025number, arhin2021surgical}. Thus, the primary value of explainability lies not in discovering novel predictors but in confirming that machine learning models learn clinically sensible reasoning. This is essential for ensuring transparency and trust in clinical applications. Simultaneously, some less obvious patterns, such as the prognostic role of “code 98” for lymph nodes (indicating the absence of pathological evaluation) or the scattered influence of rare histologic subtypes, illustrate how XAI methods can also uncover subtle, stage-dependent signals that may warrant further clinical investigation.

Overall, the findings of this study show that stage-specific modeling, combined with explainability methods, provides more accurate and clinically meaningful survival predictions. Together, these aspects strengthen the role of explainable machine learning as a trustworthy decision-support tool in oncology research. Nevertheless, as emphasized by Salih et al.~\cite{salih2025perspective}, interpretability methods such as SHAP and LIME are based on mathematical approximations; if misinterpreted or applied without awareness of their assumptions, they may lead to errors and unintended results. Therefore, the use of explainability in medicine should always be accompanied by human oversight and clinical judgment to ensure safe and responsible deployment of AI. Finally, as this study was based on SEER data, further validation in independent and more diverse cohorts is required to confirm the generalizability of the findings.

\section{Future work}Incorporating mitigation strategies and alternative XAI methods in prospective workflows will provide several benefits in clinical settings. First, validating SHAP and LIME outputs with domain experts and using multiple interpretability techniques would reduce the risk of misleading explanations, ensuring that clinicians receive reliable insights. This is critical in high-stakes decisions, such as cancer treatment planning. Second, introducing counterfactual explanations can offer actionable guidance by showing what minimal changes in patient characteristics could improve survival predictions and support personalized interventions. Similarly, gradient-based methods, such as Integrated Gradients, provide more stable and reproducible feature attributions, which can enhance confidence in model outputs. Together, these improvements would make the models more transparent and clinically actionable, fostering trust and facilitating safe integration into real-world oncology workflows.

\section*{Acknowledgment}
The authors thank the SEER program for providing access to the datasets. The authors gratefully acknowledge Prof. Ahmad P. Tafti, PhD, FAMIA, at the University of Pittsburgh, for his support and comments on this research.

\bibliographystyle{plainnat}
\bibliography{references}

\end{document}